\documentclass{article}
\usepackage{bm}
\usepackage{natbib}
\usepackage{microtype}
\usepackage{graphicx}
\usepackage{subfigure}
\usepackage{booktabs} 

\usepackage[accepted]{icml2024}

\usepackage{amsmath}
\usepackage{amssymb}
\usepackage{mathtools}
\usepackage{amsthm}

\definecolor{CColor}{rgb}{0.01,0.31,0.59}
\definecolor{GGray}{rgb}{0.80,0.90,1}
\definecolor{Shady}{rgb}{0.9,0.9,0.9}
\definecolor{kaistblue}{RGB}{20,135,200}
\definecolor{kaistdarkblue}{RGB}{0,65,145}
\definecolor{urbanablue}{RGB}{19,41,75}
\definecolor{urbanaorange}{RGB}{232,74,39}
\definecolor{drp}{rgb}{0.53,0.15,0.34}
\usepackage[plainpages=false,pdfpagelabels,colorlinks=true,linkcolor=CColor,citecolor=CColor,urlcolor=CColor]{hyperref}
\usepackage{url}
\usepackage{pifont}
\usepackage{subfigure}
\usepackage{microtype}
\usepackage{graphicx}
\usepackage{subfigure}
\usepackage[utf8]{inputenc} 
\usepackage[T1]{fontenc}    
\usepackage{url}            
\usepackage{booktabs}       
\usepackage{amsfonts}       
\usepackage{nicefrac}       
\usepackage{microtype}      
\usepackage{xcolor}         

\usepackage{adjustbox}
\usepackage{threeparttable}
\usepackage{multirow}
\usepackage{tcolorbox}
\usepackage{tikz}

\usepackage[capitalize,noabbrev]{cleveref}

\theoremstyle{plain}

\theoremstyle{definition}

\theoremstyle{remark}

\usepackage[textsize=tiny]{todonotes}

\icmltitlerunning{ }

\begin{document}

\twocolumn[
\icmltitle{Junk DNA Hypothesis: Pruning Small Pre-Trained Weights \textit{Irreversibly} and \textit{Monotonically} Impairs ``Difficult" Downstream Tasks in LLMs}





\icmlsetsymbol{equal}{*}
\begin{icmlauthorlist}
\icmlauthor{Lu Yin}{equal,Tue,Surrey} 
\icmlauthor{Ajay Jaiswal}{equal,Texas}
\icmlauthor{Shiwei Liu}{Tue,Oxford}
\icmlauthor{Souvik Kundu}{Intel}
\icmlauthor{Zhangyang Wang}{Texas}
\end{icmlauthorlist}

\icmlaffiliation{Surrey}{University of Surrey}
\icmlaffiliation{Tue}{Eindhoven University of \mbox{Technology}}
\icmlaffiliation{Texas}{University of Texas at Austin} 
\icmlaffiliation{Intel}{Intel Labs} 
\icmlaffiliation{Oxford}{University of Oxford}

\icmlcorrespondingauthor{Lu Yin}{l.yin@surrey.ac.uk}
\icmlcorrespondingauthor{Ajay Jaiswal}{ajayjaiswal@utexas.edu}
\icmlcorrespondingauthor{Zhangyang Wang}{atlaswang@utexas.edu}
\icmlkeywords{Machine Learning, ICML}

\vskip 0.3in
]



\printAffiliationsAndNotice{\icmlEqualContribution} 

\begin{abstract}
We present \textit{Junk DNA Hypothesis} by adopting a novel \textit{task-centric} angle for the pre-trained weights of large language models (LLMs). It has been believed that weights in LLMs contain significant redundancy, leading to the conception that a considerable chunk of the parameters can be removed by \textit{pruning} without compromising performance. Contrary to this belief, this paper presents a \textit{counter-argument}: small-magnitude weights of pre-trained model weights encode vital knowledge 
essential for tackling difficult downstream tasks - manifested as the \underline{monotonic relationship} between the performance drop of downstream tasks across the difficulty spectrum, as we prune more pre-trained weights by magnitude.
 Moreover, we reveal that these seemingly inconsequential weights can result in \underline{irreparable loss} of knowledge and performance degradation in difficult tasks, even when downstream
continual training is allowed. Interestingly, our evaluations show that the other popular compression, namely \textit{quantization} \textbf{fail} to exhibit similar ``monotonic" effect and does not as convincingly disentangle this task-difficulty information. To study formally, we introduce several quantifiable metrics to \textit{gauge the downstream task difficulty}: \ding{182} within the same task category, and \ding{183} across different task categories. Our extensive experiments substantiate the Junk DNA Hypothesis across a diverse range of model sizes, tasks, datasets, and even pruning methods. Codes are available at \url{https://github.com/VITA-Group/Junk_DNA_Hypothesis.git}.

\end{abstract}
\vspace{-0.5em}
\section{Introduction}

A prevailing belief suggests that neural networks contain \textbf{significant and non-significant components} across their parameters that record necessary or unnecessary expertise respectively to handle a task, which forms the inherent basis of numerous model compression techniques \cite{han2015deep}. Moreover, as model sizes continue to expand, the volume of redundant parameters is poised to escalate ~\citep{liu2022unreasonable,kim2021bert}. This principle extends its effectiveness even to billion-level Large Language Models (LLMs) ~\citep{jaiswal2023emergence,frantar2023sparsegpt,hu2021lora,lin2023awq}. With the negligible loss in performance stemming from the absence of non-significant components, a widely held belief has taken root: these non-significant components are essentially superfluous components that make a scant contribution to the model's functionality. Yet we pause and pose a question: \textit{Could there be crucial facets overlooked in the context of whether these non-significant components are truly inconsequential artifacts for large-scale models?}

This paper addresses the aforementioned query by employing \textbf{two} popular weight compression methods namely, \textit{pruning}, and \textit{quantization}, to \textit{concretely discern ``what non-significant components are in pre-trained LLM weights, what they do, and what is the best way to disentangle them"}. At this point, it is important to clarify that this paper \textit{neither aims to be an LLM compression exposition} nor to re-justify the commonsense that large pre-trained models are compressible - but rather, to use existing popular compression techniques as \textit{quantitative and easily controllable tool} to probe and comprehend the existence and the functional role of non-significant weight components. The crux of our research lies in adopting a novel task-centric viewpoint towards pre-trained weights. In other words, we investigate how the non-significant weights fraction and their embodied information \textit{correlate with model's ability to successfully perform downstream tasks of different complexity levels}.




Our extensive experiment-based study disrupts conventional assumptions by providing three key observations: \ding{182}  the assumed non-significant weight components play a pivotal role in handling challenging tasks, \ding{183} element-wise pruning simply by pre-trained weight magnitudes, despite being a naive criterion, is a reliable indicator to \textit{capture the demarcation of task difficulty} encoded within significant and non-significant components - experimentally shown as ``monotonically impairing" downstream task performance from difficult to simple, as more are pruned by magnitude. \ding{184} Meanwhile, quantization fail to display such monotonic effect and do not as convincingly disentangle this task-difficulty information. Drawing a parallel with biological insights\footnote{
Approximately 2\% of human genome encodes proteins, leaving the remaining portion seemingly superfluous~\citep{carey2015junk}. This non-coding section of the genome has earned the moniker ``Junk DNA''~\citep{Ohno1972SoM}, positing that large genomes would inevitably harbor non-coding sequences, passively accumulated over millennia, devoid of any protein-coding capacity. Yet over the past decade, it has become evident that at least some of these seemingly extraneous DNAs play essential roles in cellular function. For example, these regions of DNA contain vital sequences that act as regulatory elements~\citep{zheng2010role}. In this paper, we borrow the ``Junk DNA" nickname to refer to those ``unsung heroes" in LLM pre-trained weights - whose functionalities have been under-recognized in previous LLM compression literature.}, we nickname our discoveries as the \textbf{Junk DNA Hypothesis}:\vspace{-0.7em} 
\begin{itemize}
    \item Contrary to common beliefs, small-magnitude weights of a pre-trained model encode vital knowledge essential for tackling difficult downstream tasks - manifested as \textbf{monotonic} relationships between the amount of pruned weights and the difficulty level of tasks that start to be impaired (Section \ref{sec:same_task} \& \ref{sec:different_task}). Interestingly though, the same monotonicity observation \textit{fail} to hold for quantization, another popular compression method.\vspace{-0.5em} 
    \item Removing these ostensibly inconsequential weights can cause \textbf{irreparable} loss of knowledge and thus performance degradation on difficult tasks. This can be observed even if downstream continual training is allowed (Section \ref{sec:irreversible_claim}).\vspace{-0.5em} 
\end{itemize}

The primary challenge in formalizing and experimentally validating this conjecture lies in providing a precise and ``continuously controllable" definition of ``task difficulty", for which we explore an extensive range of options:
\begin{itemize}\vspace{-0.7em}
    \item \textbf{Within the Same Task Category}: To define task difficulty within the same task, we formulate the following \textbf{four} settings: \ding{182} vary the adequacy of target domain data~\citep{liu2019roberta} (e.g., from few-shot to full-shot); \ding{183} availability of option counts in multiple-choice question answering (QA) setting; \ding{184} influence of external information availability with varying context length for retrieval-augmented QA~\citep{ram2023context}; \ding{185} varying k-shot in-context demonstration examples to assist multiple-choice QA \cite{mosbach2023few}.\vspace{-0.5em} 
    
    \item \textbf{Across Diverse Task Categories}: To define task difficulty across different tasks, we formulate the following \textbf{two} settings: \ding{182} we utilize the gap between the best human performance, and the target LLM model's performance on a specific task (normalized by human performance), as an indicator of complexity ``sensed" by LLM for that specific task; \ding{183} we dive in comparing between  two QA tasks: factoid-based QA which involve generating precise facts about entities, versus multiple-choice QA setting where models have to choose from a set of provided answer options.\vspace{-0.5em}    
\end{itemize}

Our extensive experiments substantiate the Junk DNA Hypothesis across a diverse range of model sizes, tasks, and datasets. While the overarching notion that··more challenging downstream tasks permit less room for pruning" may not come as a surprise, our study unveils \textbf{several more subtle, often unexpected findings}: 
\begin{itemize}\vspace{-0.5em}
    \item Moving beyond the nebulous distinction between simple and complex tasks, the various conceptions of task difficulty \textbf{uniquely} defined above by us, both within and across tasks, appear to align closely with the behavior of pruning fragility. This suggests a practical desire to estimate the task-dependent achievable degree of LLM sparsity. In certain tasks, even a modest reduction in low-magnitude pre-trained weights (e.g., 10-20\%) results in a significant drop in accuracy, underscoring their pivotal role in handling intricate tasks.\vspace{-0.3em}
    
    \item Unlike pruning, we found that quantization fail to effectively capture the same \underline{``monotonically}" impairing effect of gradual compression when task complexity varied from difficult to simple. 
    
   \item Moreover, we confirm that for difficult tasks, the essential knowledge resides within the pre-trained weight values and uninformed pruning can do \underline{``irreversible"} damage for challenging tasks, i.e., the ``damage" caused by overly pruning pre-trained weights cannot be un-done by continual downstream fine-tuning.\vspace{-0.3em} 
    
    \item Junk DNA Hypothesis holds true when transitioning from unstructured to structured N:M pruning, and even extends to other weight importance-based LLMs pruning techniques (SparseGPT, Wanda). Interestingly, we found the aforementioned methods tend to benefit simple tasks more compared to challenging tasks.\vspace{-0.3em}
\end{itemize}

\begin{table*}
    \centering
     \resizebox{0.99\textwidth}{!}{\begin{tabular}{l|cccccc}
         \toprule
         \textbf{Task Setting} & \textbf{Dataset} & \textbf{Model} & \textbf{Learning} & \textbf{Pruning} & \textbf{Difficulty Variation} & \textbf{Across Task}\\
         \midrule
         \vspace{0.1em}
         \textbf{Setting 1}: Data Adequacy & GLUE benchmark & RoBERTa-Base & Fine-tuning & Magnitude, N:M & 6 & No\\
         \vspace{0.1em}
         \textbf{Setting 2}: MCQ Option Count & MMLU & Vicuna-7B & None & Magnitude, SparseGPT, Wanda, N:M & 3 & No\\
         \vspace{0.1em}
         \textbf{Setting 3}: Context Length & TriviaQA & Vicuna-7B & None & Magnitude, SparseGPT, Wanda, N:M & 5 & No\\
         \vspace{0.1em}
         \textbf{Setting 4}: k-shot examples & MMLU & Vicuna-7B & Few-shot & Magnitude, SparseGPT, Wanda, N:M & 4 & No \\
         \vspace{0.1em}
         \textbf{Setting 5}: Human-Centric & PIQA, HellaSwag, OpenBookQA & Vicuna-13B & None&  Magnitude, SparseGPT, Wanda, N:M & 3 & Yes\\
         \vspace{0.1em}
         \textbf{Setting 6}: Factoid v.s. MCQ  & FreebaseQA, MMLU & Vicuna-7B & None & Magnitude, SparseGPT, Wanda & 2 & Yes\\
         
         \bottomrule
    \end{tabular}}
    \vspace{-0.5em}
    \caption{Our experimental task settings to quantitatively control task difficulty within tasks and across tasks.}
    \vspace{-1em}
    \label{tab:task_setting}
\end{table*}

\vspace{-0.3em}
\section{Experimental Settings}
In this section, we provide a summary of our experimental settings used to validate Junk DNA hypothesis. Table \ref{tab:task_setting} provides details on dataset, model configuration, learning setting, pruning techniques, and difficulty spectrum for various task settings used to quantitatively control task difficulty with and across tasks. We {first} experiment with popular pruning methods (allow multiple data points to validate our argument): One-shot Magnitude, SparseGPT, and Wanda to comprehensively validate the \textbf{monotonic impairment} claim on our curated difficulty spectrum across various task settings. Next, in Section \ref{sec:junk_dna_compression_type}, we will draw parallel between pruning and quantization to illustrate how the later fails to capture this monotonic trend, thus unable to facilitate demarcation across task difficulty.

\textbf{Sparsity:} We consider two types of sparsities: (1) \textit{Unstructured Sparsity}:  individual weights in the model are zeroed out independently, leading to irregular zero patterns~\citep{lecun1990optimal,han2015deep}; (2) \textit{Structured \textit{N:M} Sparsity}: a fine-grained sparsity pattern in which only \textit{N} weights are non-zero for every continuous \textit{M} weights~\citep{nvidia2020,zhou2021learning}. We report the results of \textit{M=}8 (\textit{N} ranges from 7 to 1). Note that we intentionally focus on one-shot magnitude pruning, to isolate the effect of small-magnitude weights as the sole ``delta" between the two models, and due to its recently observed promise to retain LLM performance \citep{jaiswal2023emergence}. We additionally include  two more SoTA LLM pruning methods: SparseGPT \cite{frantar2023sparsegpt}, and Wanda \cite{sun2023simple} to confirm whether the validity of our hypothesis can generalize to other weight importance criteria as well.

\textbf{Dataset:} We explore a wide variety of datasets while defining task difficulty to convincingly establish the validity of Junk DNA Hypothesis. For our experiments, we use GLUE benchmark \cite{wang2018glue}, MMLU  benchmark covering 50+ subjects \cite{hendrycks2020measuring}, TriviaQA \cite{joshi2017triviaqa}, PIQA \cite{bisk2020piqa}, HellaSwag \cite{zellers2019hellaswag}, OpenBookQA \cite{mihaylov2018can}, and FreebaseQA \cite{jiang-etal-2019-freebaseqa}. More information about the aforementioned datasets is available in Appendix \ref{appendix:dataset_details}.

\textbf{Normalized Performance Drop:} Model performance depends on task category and difficulty level. To fairly illustrate the impact of compression across task category, task difficulty, and pruning techniques, our experiments present normalized performance drop to facilitate ease in perceiving conclusion. Let $f_d^t$ and $f_p^t$ be dense and pruned model performance on task $t$, normalized performance drop is calculated as $(f_p^t - f_d^t)/f_d^t$. 

To quantitatively capture the \textbf{monotonic impairment and trend} across the task difficulty spectrum, we make use of two popular statistical measures:

\textbf{1. Spearman's rank correlation:} It assesses how well the ranking relationship between two variables (X \& Y) can be described using a monotonic function. A Spearman correlation of zero indicates that there is no tendency for Y to either increase or decrease when X increases.

\textbf{2. Theil–Sen estimator:} It is a robust estimation of linear slopes (treated as a surrogate of ``how fast Y changes with X") from many (potentially noisy) sample points (X \& Y). It works by choosing the median of the slopes of all possible lines that go through any pair of points. Higher slopes by Theil–Sen estimator indicate faster changes of Y w.r.t. X.

\section{How Pruning Illicit Monotonic Impairment Across Difficulty Spectrum?}
\label{sec:monotonic_impair}
\subsection{Case study of difficulty spectrum within same task}
\label{sec:same_task}
\subsubsection{Task Difficulty Setting 1: \textit{Varying the Adequacy of Target Domain Data}}
\label{sec:pruning}
\textbf{Rationale:} The difficulty of learning a task is commonly thought to be influenced by the number of available training examples: fewer data points typically imply more challenges to learn well. To quantitatively control task difficulty within a single task, we manually manipulate the \textit{volume of data used for fine-tuning} by randomly sampling certain ratios of data from the target domain dataset. This allows us to \textit{disentangle task difficulty from the task type}. 

\begin{figure}[h]
\centering
    \includegraphics[width=0.95\columnwidth]{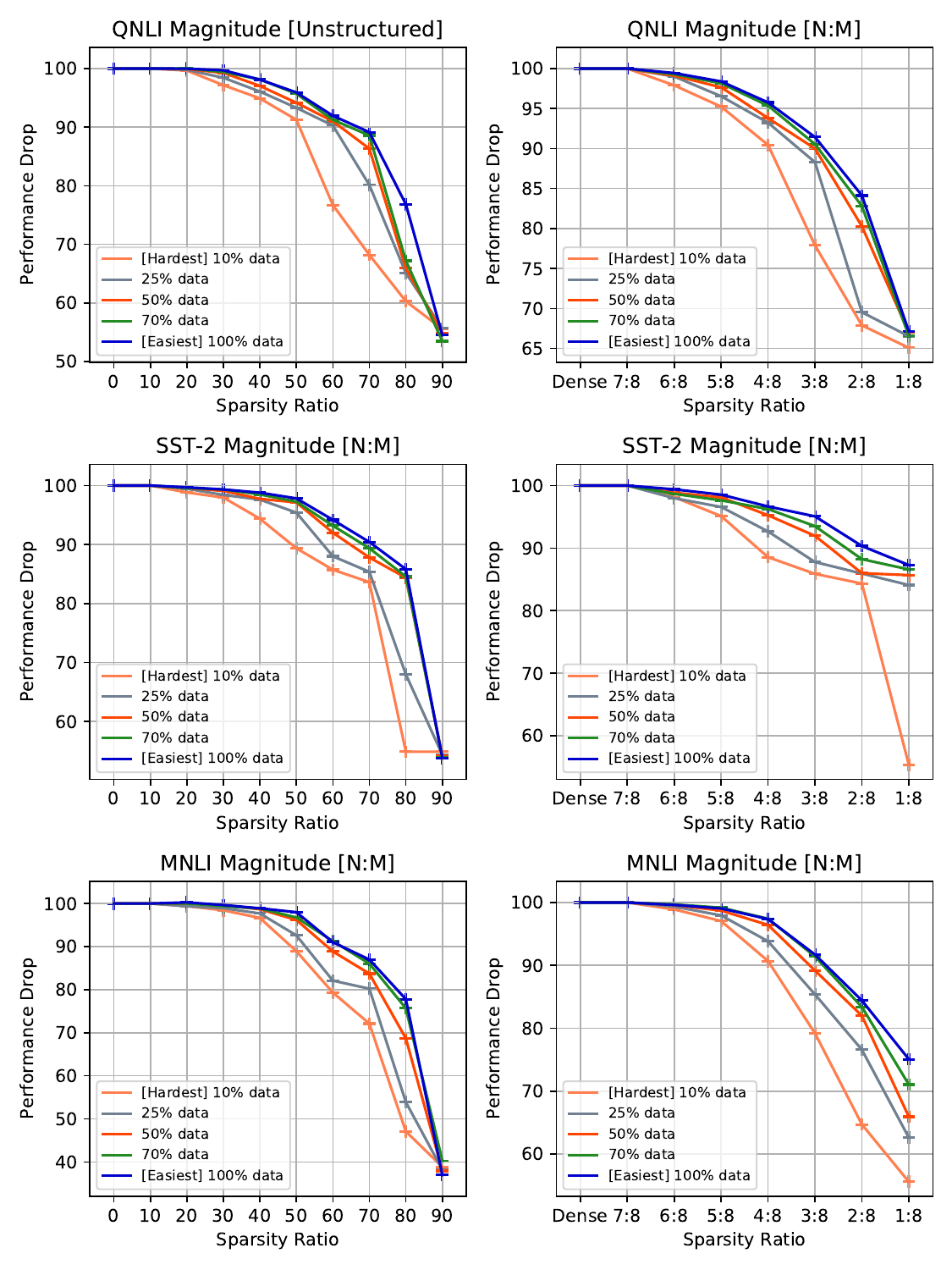}
\vspace{-1em}
\caption{\textbf{Task Difficulty Setting 1: Varying target domain data adequacy}: Dense Transfer vs. Sparse Transfer using RoBERTa-Base on various downstream tasks. Task difficulty is measured by the training data volume. }
\label{fig:task_setting1}
\vspace{-1.5em}
\end{figure}

\textbf{Method:} To examine the influence of small-magnitude weights, we conduct a comparative analysis between two models: one starting from the pre-trained model with small-magnitude weights, and the other without. The former is commonly referred to as task-specific fine-tuning on downstream tasks, denoted as \textbf{Dense Transfer} in this paper. The latter model, named \textbf{Sparse Transfer}, differs from dense transfer in the way that we first perform magnitude pruning on the pre-trained model, creating a sparse model. We then fine-tune on downstream tasks while keeping the sparse mask fixed. Figure \ref{fig:task_setting1} summarizes our results for this setting across unstructured and structured sparsity patterns.

\begin{figure}[h]
\centering
    \includegraphics[width=0.95\columnwidth]{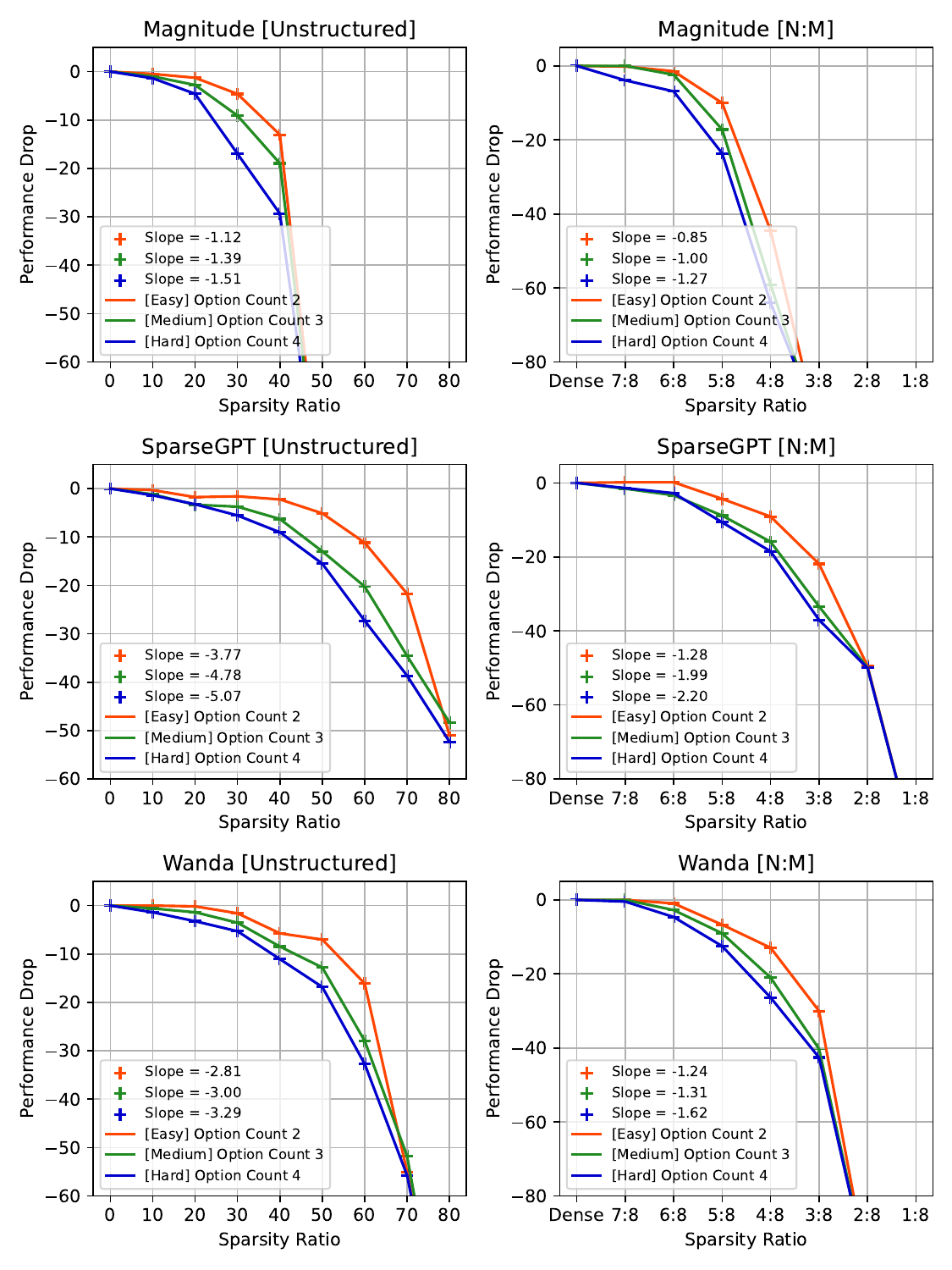}
\vspace{-1em}
\caption{\textbf{Task Difficulty Setting 2: Varying option count in multiple-choice QA}. Dense v.s. Sparse subnetwork performance of Vicuna-7B using MMLU. Task difficulty is measured by the number of available choices in a multiple-choice QA setting. }

\label{fig:task_setting2}
\vspace{-1.5em}
\end{figure}

\subsubsection{Task Difficulty Setting 2: \textit{Varying the Option Count in Multiple-choice QA Setting}}

\textbf{Rationale:} This setting proposed in our work, extends the idea that given a task of multiple-choice question answering with a fixed dataset, the difficulty of the task is proportional to the number of available options to select the correct answer. To quantitatively control task difficulty, we manually manipulate the {option count} \textit{for each question from }[$2-4$] which provide a random guess success rate from 50\% (2 options) to 25\% (4 options). This setting uniquely allows us to control the task difficult for a given task. 

\textbf{Method:} To understand the influence of small-magnitude weights, in this setting, we exploit {in-context learning} setting with a natural prompting approach to present the question and answer options to the LLMs jointly, and have it output the symbol (\emph{e.g.}, ``A") associated with its chosen answer option. We fixed 5 in-context question-answer example pairs while . Here, we attempt to understand how the pruned model can effectively reason and successfully associate the correct answer among the given answer options with symbols that represent them, using knowledge remaining within them with increasing difficulty by varying option counts. Figure \ref{fig:task_setting2} summarizes our results for this setting across unstructured and structured sparsity patterns.

\begin{figure}[h]
    \centering
    \includegraphics[width=0.99\columnwidth, trim=0 3em 0 0]{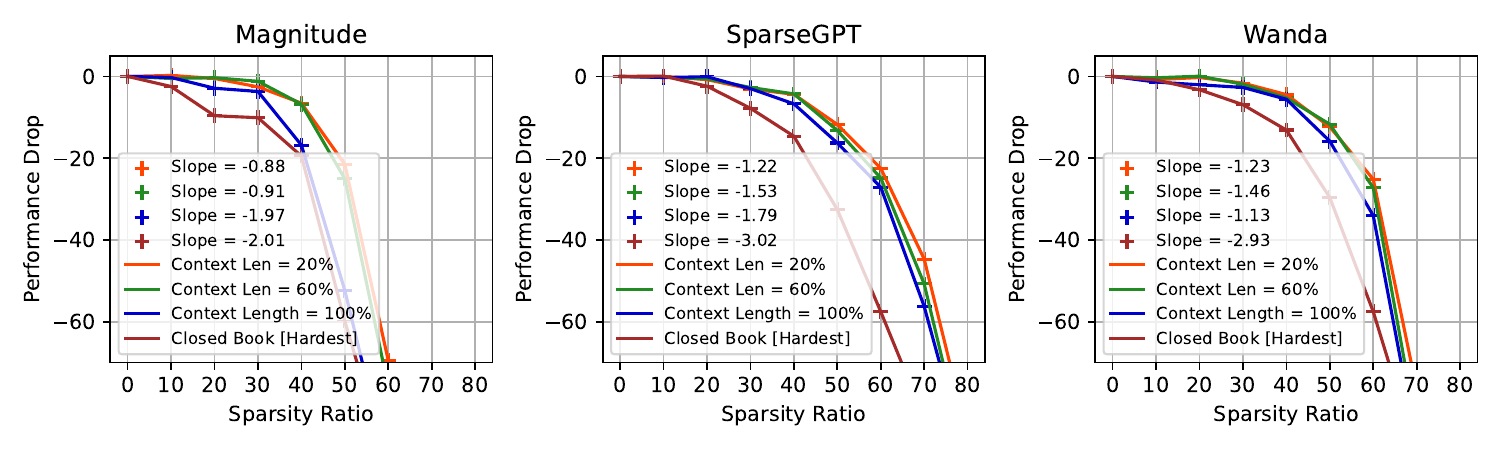}
    \caption{\textbf{Task Difficulty Setting 3:Varying context length in Retrival-Augmented QA}. Dense v.s. Sparse subnetwork performance of Vicuna-7B pruned using TrivaQA Benchmark. Task difficulty is measured by the number of tokens provided in context.}
    \label{fig:task_setting3}
    \vspace{-1em}
\end{figure}

\begin{figure}[h]
    \centering
    \includegraphics[width=0.99\columnwidth, trim=0 3em 0 0]{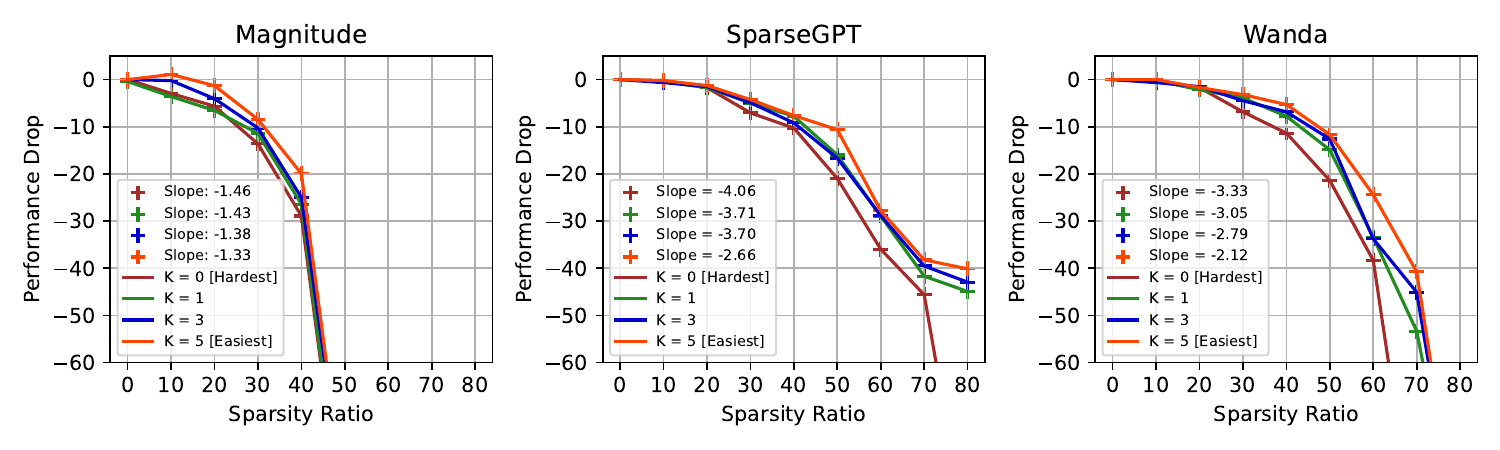}
    \caption{\textbf{Task Difficulty Setting 4:Few-shot In-context Learning}. Dense v.s. Sparse subnetwork performance of Vicuna-7B pruned using MMLU Benchmark. Task difficulty is measured by the number of $K$ shot in-context demonstration examples provided to assist multiple-choice QA.}
    \label{fig:task_setting4}
    \vspace{-2em}
\end{figure}

\subsubsection{Task Difficulty Setting 3: \textit{Varying context length for Retrieval-Augmented QA}}
\textbf{Rationale:} In this task setting, we exploit a highly industry-relevant task of Retrieval-Augmented QA in which given a context, LLMs are prompted to answer questions based on the context. To define a quantitative measure of difficulty for this task, we propose to \textit{vary the context length ensuring that the correct answer still resides within the provided context}. Retrieval-augmented QA requires LLMs to possess sufficient ability to synthesize long in-context knowledge provided within input prompts, and locate and retrieve correct answers within it. Our novel strategy to \textbf{control token counts} within context helps us to define a smooth relation with the task difficulty. 

\textbf{Method:} Our evaluation system using TriviaQA includes two high-level components: \ding{182} \textit{document selection}, selecting the set of documents upon which to condition; and \ding{183} \textit{document reading}, determining how to incorporate the selected documents into the LLM answer generation process, which requires extracting correct answer phrases from conditioned documents. Given the ground truth, we select x\% of tokens around it in the context document from the document selection step, to ensure that the context is complete (\emph{e.g.,} no split sentences). We evaluate the performance of pruned LLMs to investigate their ability to generate answers using the provided x\% tokens. We additionally added a closed-book \cite{ram2023context} setting at the end of the difficult spectrum (highest) because in this setting, LLMs are not provided with any external information and need to rely on their own internal knowledge ingested during pretraining \cite{jaiswal2023compressing}.

\subsubsection{Task Difficulty Setting 4: \textit{Varying number of k-shot examples for in-context learning}}
\textbf{Rationale:} Modern-day LLMs have shown a unique ability for out-of-domain generalization using in-context learning (ICL). ICL adapts a model to a task by conditioning it with few-shot demonstrations specified purely via text interaction with the model and it has been observed to significantly improve performance on the downstream application \cite{brown2020language, mosbach2023few}. In this task setting, we propose to \textit{control the amount of example demonstrations to quantitatively measure the difficulty of the task}. Based on our experimental finding that providing more demonstrations leads to better performance of the dense model, this setting defines the difficulty spectrum directly on the number of example demonstrations provided to the pruned model during the conversation using natural language prompts.

\textbf{Method:} Similar to task setting 2, we use in-context learning to present $k$ example demonstration to the compressed model. For our evaluation, we use MMLU benchmark, and use $k \in \{0, 1, 3, 5\}$ examples from the validation set to control the task difficulty where 0 indicates no example was shown to the model (hardest) while 5 indicate the model generated answer observing 5 examples (easiest). Figure \ref{fig:task_setting4} summarizes our results for this setting.

\subsubsection{Main Results: Validating the Junk DNA }

\textbf{\ding{182} Removal of small-magnitude weights is viable to some extent for easier tasks in comparison to difficult tasks:} In all four task setting's results (Figure \ref{fig:task_setting1}, \ref{fig:task_setting2}, \ref{fig:task_setting3}, and \ref{fig:task_setting4}), we find that it is feasible to discard around 10-15\% of small-magnitude at once without significantly compromising the performance in unstructured sparsity. This indicates that \textit{for simple tasks, the knowledge encoded in high-magnitude weights is sufficient to handle the task}. For instance, in task setting 4 (Figure \ref{fig:task_setting4}a), we found that for the easiest task (with 5 ICL demonstration), we can remove small weights up to 14\% without any drop in performance in comparison with dense counterpart, which is \textbf{not true} for the hardest task (with 0 ICL demonstration).

\textbf{\ding{183} Eliminating small weight presents a monotonic impairment relationship with increasing task difficulty:}
Across all our experiments, we found a smooth monotonic impairment relationship between removing small weights and the difficulty of the task. As we gradually move to the hard end of the difficulty spectrum, the impact of pruning small weights keeps increasing which shows that the “useless” small weights are imperative to encode crucial knowledge necessary to solve more challenging downstream tasks. The \textit{monotonically increasing theil–sen estimator slope w.r.t. task difficulty provides a quantitative estimate to measure “monotonically impairing" caused by pruning} low-magnitude weights.

\textbf{\ding{184} Junk DNA persists in both N:M sparsity and unstructured sparsity, and beyond the magnitude criteria}
Both N:M sparsity and unstructured sparsity yield similar observations, indicating the presence of Junk DNA in both settings. We further experiment with two other popular LLM-pruning methods: SparseGPT and Wanda which consider weight importance as a criterion for pruning. Our observations can be summarized as: (1) both these methods (including their N:M setting) corroborate with JunkDNA Hypothesis, (2) JunkDNA can be viewed as a benchmark to develop better pruning algorithms which can minimize the performance gap between easy and hard task, (3) interesting, we observe that these \textit{new sophisticated weight importance based pruning methods tend to benefit simple task more in comparison to challenging tasks}, indicating a need to carefully evaluate true merits of modern compression methods. 

\subsection{Case study of difficulty spectrum across tasks}
\label{sec:different_task}
\subsubsection{Task Difficulty Setting 5: \textit{Estimating LLM-facing Task Difficulty by Normalized Human-LLM Performance Gap}}

\textbf{Rationale and Method:} We propose a method to gauge complexity by juxtaposing the performance of deep learning models with that of human counterparts. Specifically, we define task difficulty as the disparity in performance between humans and models, normalized by human performance. A more pronounced positive performance gap (for instance, where humans outperform the machine to a greater extent) would signify a higher level of difficulty for the model in handling the given task. Conversely, in cases where the machine outperforms humans, a larger gap indicates an easier task. The resulting assessment of across-task difficulty is outlined in Table~\ref{tab:human_measure_task}. Specifically, we choose a range of downstream tasks, including PIQA, HellaSwag, OpenBookQA and our experiments are conducted using Vicuna-13B.

\begin{figure}[h]
\centering
    \includegraphics[width=0.9\columnwidth, trim = 0 0 0 2em]{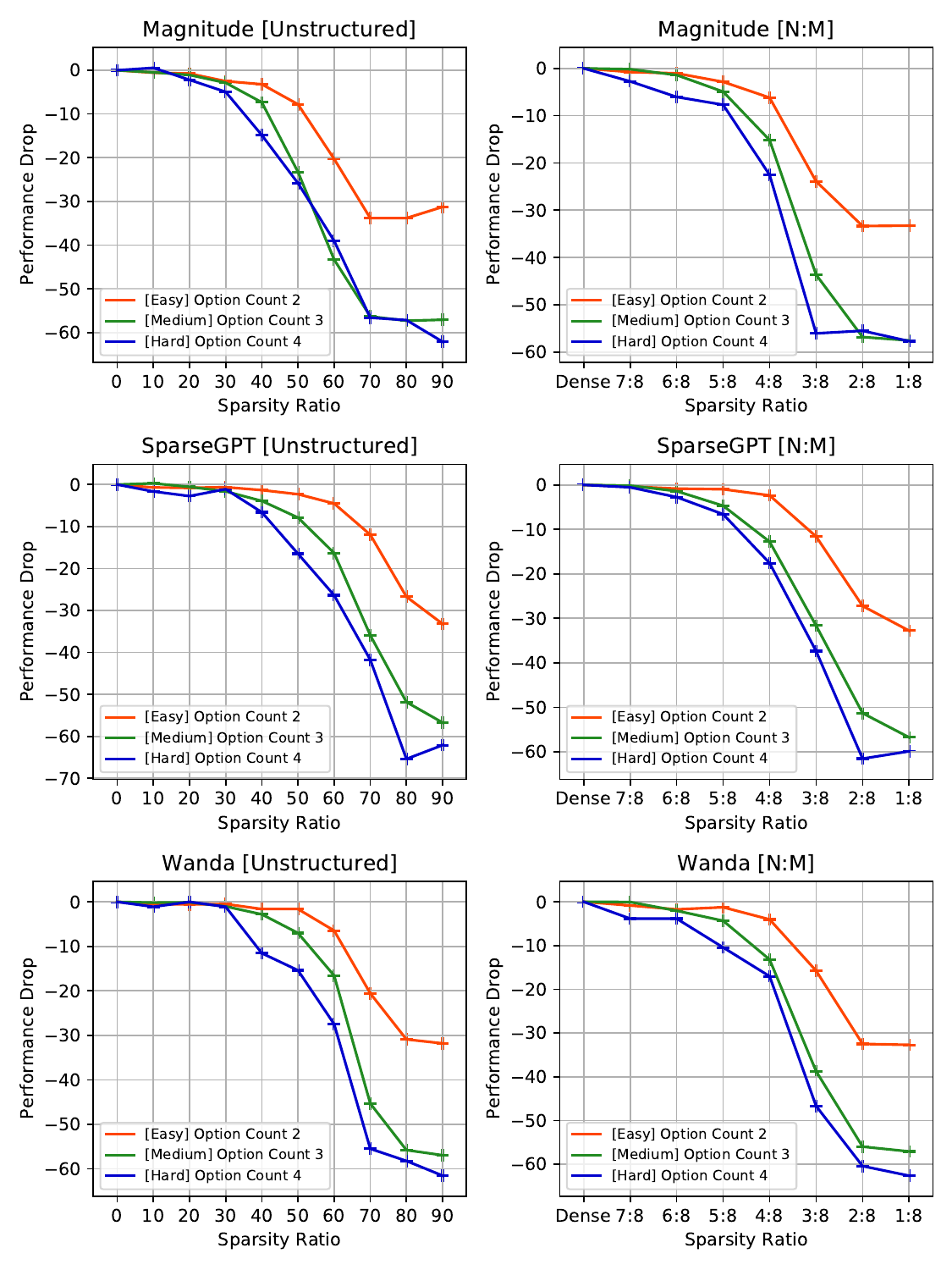}
\vspace{-1em}
\caption{\textbf{Across-Task Difficulty via Normalized Human-LLM Performance Gap:} Dense v.s. Sparse subnetwork performance of Vicuna-13B. Task difficulty is measured by  Human-LLM Performance gap \textbf{normalized} by the dense performance.}
\label{fig:task_setting5}
\vspace{-1.5em}
\end{figure}

\subsubsection{Task Difficulty Setting 6: \textit{Factoid-based v.s. Multiple-choice QA}}
\textbf{Rationale and Method:} In this setting, we compare two popular QA settings: Factoid-based QA and Multiple-Choice QA.  A typical Factoid-QA task aims to answer natural language questions using facts,  i.e., entities or attributes of knowledge ingested within them during pre-training. On the other hand, multiple-choice QA setting present the question and answer options to the LLMs jointly, and have it output the symbol (\emph{e.g.}, ``A") associated with its chosen answer option. Unlike multiple-choice QA, factoid-based QA requires  LLMs’ ability to answer natural language questions using facts using its own knowledge and possess the ability to generate full answers unlike output a symbol, which makes it inherently a more challenging task.

\begin{figure}[h]
    \centering
    \includegraphics[width=0.99\columnwidth, trim=0 3em 0 0]{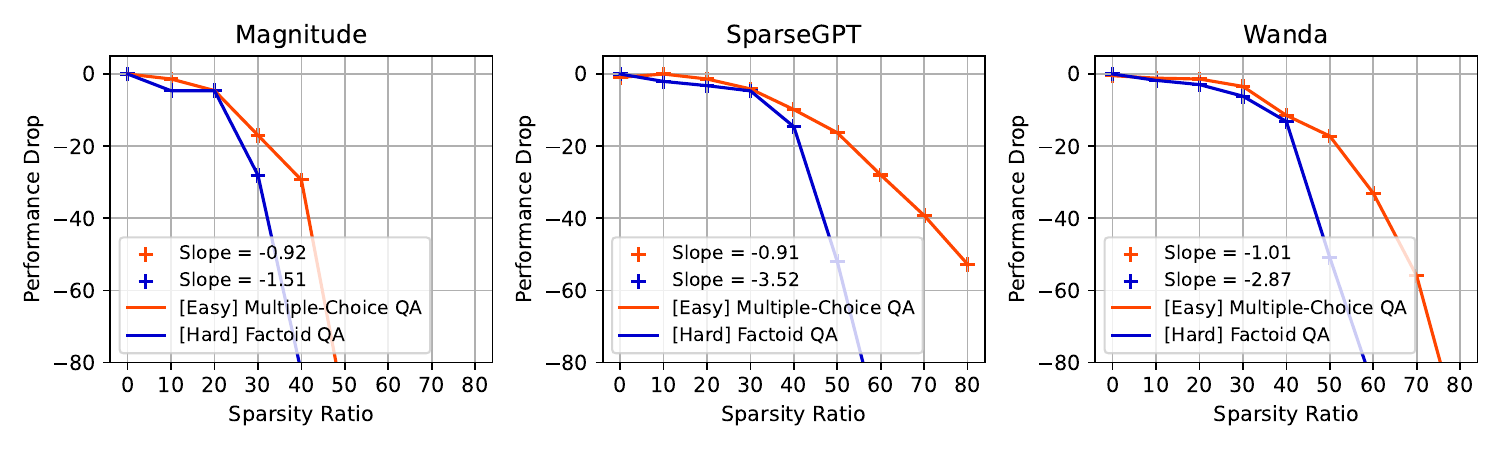}
    \caption{\textbf{Across-Task Difficulty for Factoid-based QA and Multiple-Choice QA:}: Dense v.s. Sparse subnetwork performance of Vicuna-7B. Task difficulty is measured by  Human-LLM Performance gap \textbf{normalized} by the dense performance.}
    \vspace{-1.5em}
    \label{fig:task_setting6}
\end{figure}

\subsubsection{Main Results: Validating the Junk DNA Hypothesis}
The findings depicted in Figure~\ref{fig:task_setting5},~\ref{fig:task_setting6} echo the conclusions drawn in Section~\ref{sec:same_task}, once more providing robust support for the validity of the \textit{Junk DNA hypothesis across a broad spectrum of task categories}. While it may be feasible to remove small-magnitude weights without significant repercussions in simpler tasks, these pre-trained small weights contain vital downstream knowledge essential for tackling challenging tasks and thus are no longer dispensable suggesting their importance -manifested as monotonic relationships between the amount of pruned weights and the difficult level of tasks that start to be impaired.

\section{Does Quantization Also Yield Monotonic Impairment Across Difficulty Spectrum?}\label{sec:junk_dna_compression_type}
\textit{Pruning} and \textit{quantization} are two popularly used remedies to reduce the overheads of billions of parameters in current LLMs. {Pruning} techniques primarily shrink network sizes by removing specific weights from the model – essentially setting them to zero. {Quantization} methods aim to quantize parameters into low bit-precision to reduce compute and memory budget. Across recently developed quantization methods for LLMs \citep{Lin2023AWQAW, Frantar2022GPTQAP, Dettmers2023SpQRAS}, we adopted GPTQ for the quantization of all linear layers (exactly similar to our pruning settings) to study the monotonic impairment induced across the task difficulty spectrum. In this section, we firstly ask: \textit{How does the ratio of compression using quantization reflect on the performance of the downstream tasks? Does quantization strength dictate any performance pattern across the difficulty spectrum of the underlying task?}  

\begin{figure}[h]
    \centering
    \includegraphics[width=0.98\columnwidth, trim = 0em 2em 0em 0]{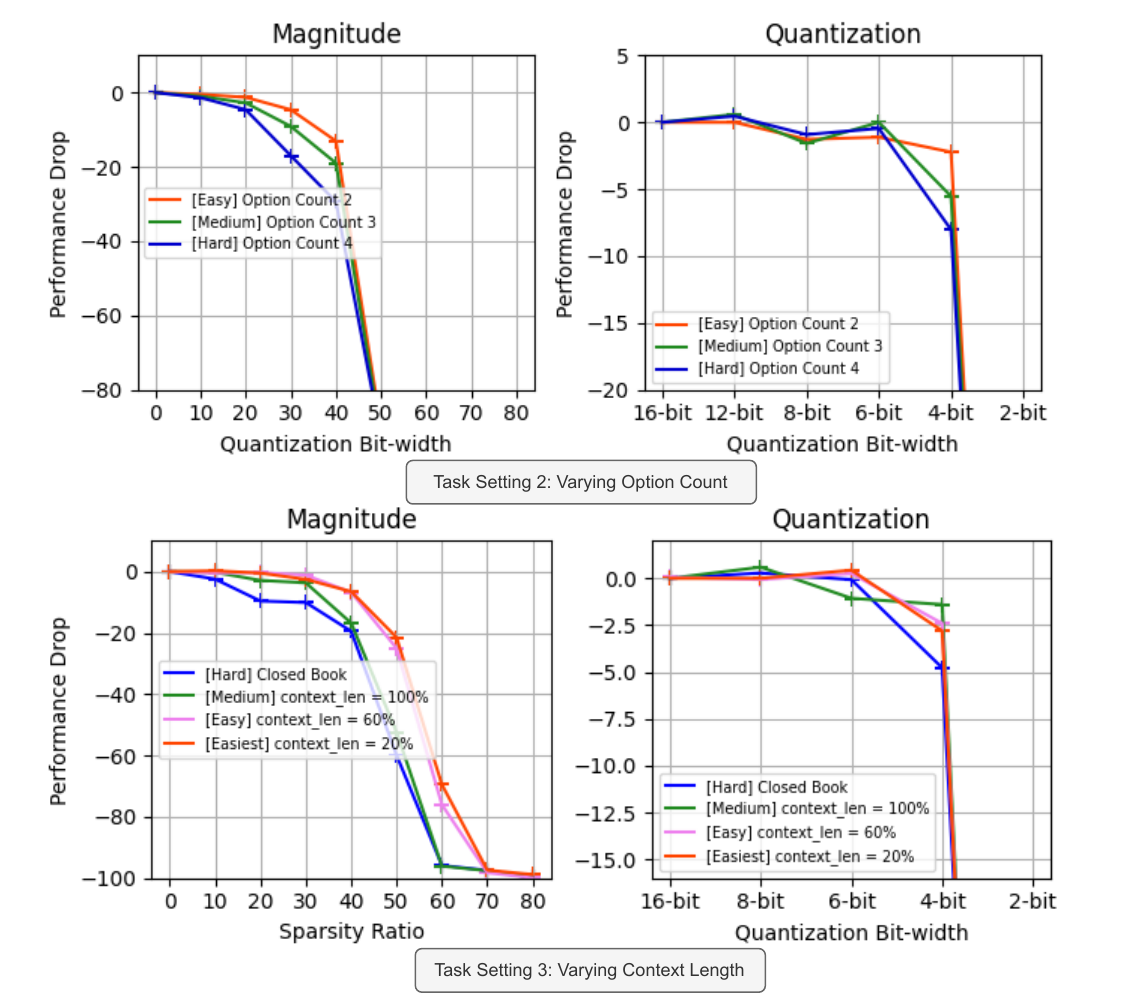}
    \caption{\textbf{How is pruning special?} Performance comparison of pruning and quantization with varying compression ratios on our task difficulty spectrum. We can observe the \textbf{monotonic impairment} of pruning across task difficulty and pruning ratio. On the contrary, quantization \textbf{fail} to capture this monotonic behavior across task difficulty and compression ratio.  
    }
    \label{fig:pruning_quant}
    \vspace{-1.7em}
\end{figure}

To this end, Figure \ref{fig:pruning_quant} presents our head-to-head comparison for one-shot low-magnitude pruning and GPTQ quantization \cite{frantar2022gptq} various compression strengths. We adopted retrieval-augmented QA (task setting 3) in which task difficulty is controlled by varying the context length (token counts) accommodating the correct answer phrase, and variable option count in multiple-choice QA (task setting 2) for our experiments.  

Based on our experimental results, we summarize: \ding{182} while pruning shows a monotonic performance degradation with increasing compression ratio, quantization does not exhibit the smooth impairing impact of compression; \ding{183} removal of low magnitude weights using pruning tends to have a monotonically increasing impairing impact if we move along the extreme end of our task difficulty spectrum, which is again not convincingly demarcated either by quantization. It is interesting to observe that quantization to a certain degree tends to benefit the overall performance but again without any clear pattern across the difficulty spectrum. 

\textbf{How Is Pruning Special?} It is interesting to see the unique ability of pruning to identify a critical subset of weights that can independently complete specific set of tasks the same way as the dense model - but not quantization. This naturally reminds us the Lottery Ticket Hypothesis (LTH) \cite{frankle2019lottery,chen2020lottery}, stating that there exists a pruned subnetwork that can match the performance of its dense counterpart; meanwhile no similar conclusion has been drawn for quantized networks at scale.  Such discrepancy seems to echo the two methods' different impairment behaviors (monotonic or not across task difficulty levels). We additionally experimented with low-rank compression \cite{yu2017compressing} (not the low-rank weight updates used in parameter-efficient tuning \cite{hu2021lora}) and found its behavior closer to pruning than quantization (Appendix \ref{appendix:lrc}). 

The classical LTH (with iterative pruning and re-training) has not yet been studied for modern-scale LLMs due to its prohibitive computational costs. Previous work in smaller models did find reusable sparse substructures that contain sufficient information to handle a few similar tasks altogether \cite{chen2020lottery,yu2019playing}. Recent attempt \cite{jaiswal2023emergence} demonstrated a similar property on LLMs for a range of tasks, using cheap one-shot pruning without re-training; but the authors also recognized the ``sweetpoint sparsity" was correlated to the difficulty of downstream tasks involved (although they did not provide a concrete way to quantify such). As a modern LLM can perform an infinite range of downstream tasks in zero- or few-shot, Junk DNA Hypothesis can be viewed as a scaled extension of LTH,  which suggests that: if the task difficulty spectrum (much broader than before) is taken into consideration, then there might not exist a single ``lottery" subset of weight, that can be removed without causing irreparable damage for the performance of \textit{all tasks}  - although one might still be able to find a ``lottery" for \textit{some (easy) tasks}.

\begin{table}
    \centering
    \resizebox{0.71\columnwidth}{!}{\begin{tabular}{ll|cc}
    \toprule
         Task Setting & Difficulty ($\downarrow$) & Pruning & Quantization\\
    \midrule
  
         Task Setting 3 &Closed Book        & 0.996 & 0.799\\
         &Context\_len = 100\% & 0.996 & 0.794\\
         &Context\_len = 60\%  & 0.975 & 0.399\\
         &Context\_len = 20\%  & 0.987 & 0.316\\
    \midrule
        Task Setting 2 & Option Count 4        & 0.979  & 0.872\\
         &Option Count 3        & 1.0    & 0.799\\
         &Option Count 2        & 1.0    & 0.666\\
    \bottomrule
    \end{tabular}}
    \vspace{-0.7em}
    \caption{Spearman’s rank correlation of pruning and quantization for task with varying difficulty levels. Unlike quantization high Spearman correlation for pruning uniquely captures monotonic impairment induced by pruning.}
    \label{tab:spearman}
    \vspace{-1.8em}
\end{table}

\begin{figure*}[h]
\centering
    \subfigure{
        \includegraphics[width=0.9\textwidth, trim=0 10em 0 0em]{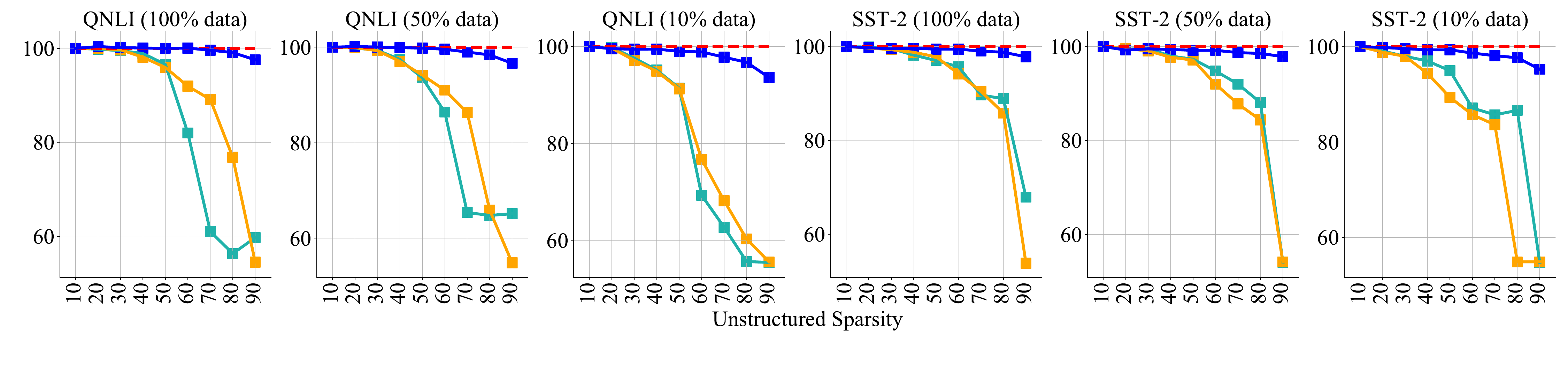}
    }
    
    \subfigure{
        \includegraphics[width=0.90\textwidth]{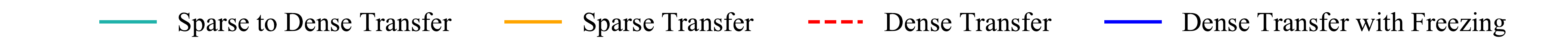}
    }
    \vspace{-1.5em}
\caption{\textbf{Varying target domain data adequacy}: Four different fine-tuning settings with RoBERTa-Base on various downstream tasks. All performance is normalized by the one of Dense Transfer.}
\label{fig:gem_within_difficulty}
\end{figure*}

\begin{figure*}
\vspace{-0.8em}
\centering
    \subfigure{
        \includegraphics[width=0.91\textwidth, trim=0 8em 0 0em]{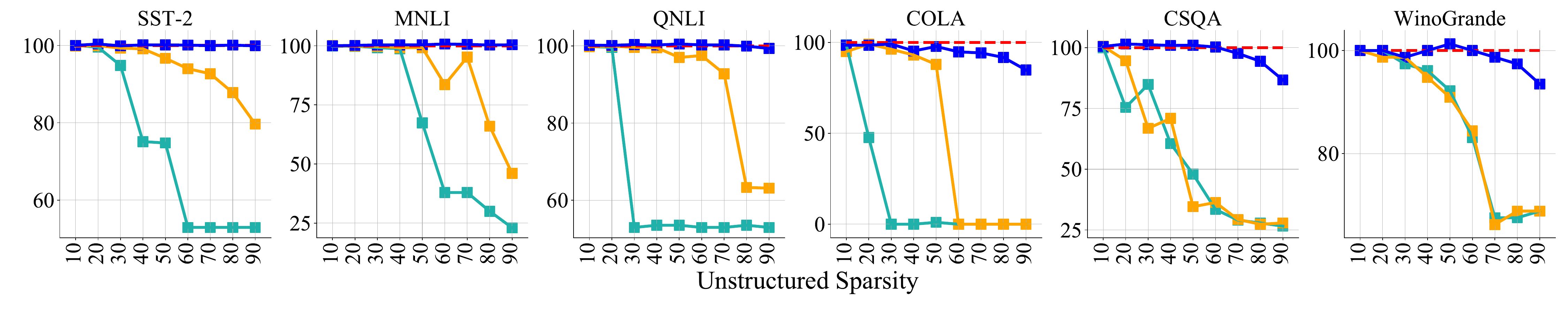}
    }

    \subfigure{
        \includegraphics[width=0.9\textwidth]{images/legend.pdf}
    }
    \vspace{-1.5em}  
\caption{\textbf{Across-Task Difficulty via Normalized Human-LLM Performance Gap}: Four different fine-tuning settings with RoBERTa-Large on various downstream tasks.  All performance is normalized by the one of Dense Transfer. }
\label{fig:gem_across_difficulty}
\vspace{-1.3em}
\end{figure*}

\section{Are Pre-trained Magnitude Values Indeed the True Gem? } 
\label{sec:irreversible_claim}
Having recognized the pivotal role of small weights in downstream adaptation, particularly in relation to task difficulty, our next objective is to delve into the foundational factors contributing to the crucial function of small weights and whether they can be recovered with fine-tuning. More specifically, we intend to validate if pruning leads to \textbf{irreparable} loss of knowledge. Our primary research inquiries are:

\textbullet\ \textit{Which one holds greater significance: the knowledge (weight-values) stored in pre-trained small-magnitude weights, or the potential to adjust these weights through fine-tuning?}

\textbullet\ \textit{Is it possible to recover the knowledge embedded in pre-trained small-magnitude weights if we prune them and allow them to grow freely during fine-tuning?}

\textbf{Method:} We explore four comparison methods: (1) \textbf{Dense Transfer}: as described in Section~\ref{sec:pruning}; (2) \textbf{Dense Transfer with (Partial) Freezing}:  a dense model where small-magnitude weights remain fixed during fine-tuning; (3) \textbf{Sparse Transfer}: as in Section~\ref{sec:pruning}; (4) \textbf{Sparse to Dense Transfer}: small-magnitude weights are initially pruned after pre-training, and subsequently during fine-tuning, those pruned weights are allowed to gradually regain non-zero values. This approach also aids in determining whether the knowledge within small-magnitude pre-trained weights is essential for performance or if their adaptability during fine-tuning takes precedence. We pick Task Setting 1 (within the same task) and Task Setting 5 (across different task), to report their performance using RoBERTa-Large, on MNLI, QNLI, SST-2, as well as CSQA and WinoGrande. The difficulty spectrum normalized by human performance for the aforementioned tasks is presented in Appendix \ref{appendix:human_performance}.

\begin{table}[h]
\centering 
\vspace{-1.5em}
\caption{Measuring the Across-Task Difficulty by the Performance Difference between humans and models (normalized by human performance): the larger (positive) margin, the more difficult for the model.}
\resizebox{0.4\textwidth}{!}{
\begin{tabular}{l|ccc}
\toprule
& PIQA& HellaSwag & OpenBookQA  \\ 
\midrule
Human  & 95.0 & 95.6 &  92.0  \\ 
\midrule
Vucuna13B & 78.35& 59.68  & 36.40 \\
\midrule
``Task Difficulty" (\%)  & [Easy] 17.53  & [Medium]   37.57  &   [Hard]  60.43   \\
\bottomrule
\end{tabular}
\label{tab:human_measure_task}
}
\vspace{-1.0em} 
\end{table}

\textbf{Results:} The outcomes for both within-task difficulty and across-task difficulty are illustrated in Figure~\ref{fig:gem_within_difficulty} and Figure~\ref{fig:gem_across_difficulty}, respectively. Below, we outline our key observations:

\textbf{\ding{182} Pre-trained small weights harbor vital downstream knowledge, beyond mere free parameters.} Across both task-difficulty metrics, it becomes evident that settings preserving the pre-trained values of small weights—namely, Dense Transfer and Dense Transfer with Freezing—achieve superior performance when compared to the other two settings. The removal of small-magnitude weights from pre-trained models results in significant performance degradation, even when we permit the pruned weights to regenerate during fine-tuning. This observation strongly bolsters the Junk DNA Hypothesis, indicating that small-magnitude weights are far from redundant; rather, they house sophisticated knowledge crucial for downstream adaptation. This knowledge proves challenging to re-gain through fine-tuning, if these initial pre-trained weights are eliminated.

\textbf{\ding{183} Freezing without updating yields commendable results.} Remarkably, on simpler tasks like SST-2, MNLI, and QNLI, maintaining an overwhelmingly large portion (90\%) of small-magnitude weights in a frozen state leads to equally impressive performance without any loss. Even on more intricate tasks such as COLA, CSQA, and WinoGrande, freezing up to 70\% of small-magnitude weights results in no discernible performance dip. This suggests that for easier tasks, the knowledge embedded in pre-trained small-magnitude weights is already more than sufficient. However, for more challenging tasks, allowing for moderate updates to all pre-trained weights remains essential.

\section{Conclusion and Future Work}
In this study, we embark on an exploration to validate the prevailing belief that deep network weights are excessively redundant, allowing for a substantial pruning of parameters without compromising performance. Our research presents a compelling counter-argument by unearthing the previously overlooked yet intricate role of small-magnitude weights, closely tied to the difficulty level of downstream tasks. We found small-magnitude pre-trained weights encode vital knowledge essential for tackling difficult downstream tasks - manifested as monotonic relationship between increasing task difficulty and impairment induced by removing small weights. Moreover, we argue that these seemingly inconsequential weights can result in irreparable loss of knowledge and performance degradation in difficult tasks, even if downstream continual training is allowed. Meanwhile, quantization fails to exhibit similar effects
and hence does not as convincingly disentangle this task-
difficulty information as pruning. As the number of parameters in deep networks continues to grow exponentially, our findings prompt the exploration of directions such as task-complexity-dependent dynamic inference and network self-slimmable properties.

\section{Impact Statements}


This study presents the "Junk DNA Hypothesis," a nuanced perspective on the role of small-magnitude weights in large language models (LLMs), challenging the traditional view of their redundancy. Our findings highlight the importance of these weights in complex tasks and suggest the need for further exploration into dynamic inference and network slimming techniques based on task complexity. This could lead to improved methods for more efficient inference, and contribute to the ``GreenAI" goal.

\section*{Acknowledgements}
This work used the Dutch national e-infrastructure with the support of the SURF Cooperative using grant no. NWO-2023.027.


\bibliography{example_paper}
\bibliographystyle{icml2024}

\newpage
\appendix
\onecolumn

\section{Related Work}
\label{appendix:related_work}
\vspace{-0.5em}

\subsection{Classical Pruning and Sparse Neural Networks}
\vspace{-0.3em}
Pruning removes specific parts of a deep neural network, such as weights, neurons, or filters. The initial purpose of pruning is retrospectively to accelerate the model at inference time (a.k.a., post-training sparsification~\citep{mozer1989using,lecun1990optimal}. Post-training sparsification has been well studied and results in various mature criteria that can be generally categorized into zero-order methods~\citep{han2015deep,gale2019state}, first-order methods~\citep{molchanov2016pruning,sanh2020movement,jiang2021towards}, and second-order methods~\citep{lecun1990optimal,hassibi1992second,dong2017learning} - the last usually achieve higher performance but  also are more expensive due to the Hessian calculation, leading to the development of many Hessian approximation approaches~\citep{zeng2018mlprune,wang2019eigendamage,singh2020woodfisher,kurtic2022optimal}. The Lottery Ticket Hypothesis (LTH)~\citep{frankle2018the} utilizes iterative magnitude pruning (IMP) to identify a subnetwork at initialization that can be re-trained independently to the original dense network's performance. Sparse training~\citep{mocanu2018scalable,2022training,mostafa2019parameter,evci2020rigging,liu2021we,yuan2021mest,yin2023dynamic, kundu2021dnr}, on the other hand, starts with a (random) sparse network and updates network connectivity during training to search for good sparse neural network without any pre-training and dense training steps.

\vspace{-0.3em}
\subsection{Sparsity in Large-Scale Models}
\vspace{-0.2em}
The advent of large-scale pre-trained models has led to the development of advanced post-training pruning methods, aiming to enhance the cost-effectiveness of these expansive models~\citep{sanh2020movement,chen2020lottery,jaiswal2023instant,zafrir2021prune,kurtic2022optimal,xu2021rethinking,lagunas2021block,zhang2022platon,frantar2021m,jaiswal2023emergence,ma2023llm}. Among them,~\citet{frantar2021m} extend second-order pruning to the BERT-level scale, enabling the pruning of blocks of weights and achieving state-of-the-art results for sparse BERT.~\citet{frantar2023sparsegpt} introduce SparseGPT for pruning large language models (LLMs) in a single shot without requiring re-training or fine-tuning. They leverage column-wise second-order pruning, and successfully remove 100B weights from OPT-175B without a significant increase in perplexity. More recently,~\citet{sun2023simple} propose a straightforward pruning method that takes both weights and activations into account, demonstrating comparable performance to~\citet{frantar2023sparsegpt}.~\citet{li2022large} reveal that activation sparsity is a prevalent phenomenon in Transformers (90\% of intermediate output), yielding another opportunity for acceleration. \citet{liu2023sparsity} introduce a large-scale SMC-Bench, indicating that state-of-the-art magnitude- and/or gradient-based sparse algorithms fall short when applied out-of-the-box to larger-scale models and a selected of complex downstream tasks. Our study is inspired by~\citet{liu2023sparsity}, but with significantly expanded experiment scales, versatile task choices, concrete task difficulty definitions, and richer insights.

\section{Across-Task Difficulty by the Performance Difference between humans and models. }
\label{appendix:human_performance}
\begin{table*}[h]
\centering
\vspace{-0.5em}  
\caption{Measuring the Across-Task Difficulty by the Performance Difference between
humans and models (normalized by human performance): the larger (positive) margin, the more difficult for the machine. Human performance is obtained from~\citet{nangia2019human}.  The more challenging task is marked in bold.}
\resizebox{0.99\textwidth}{!}{
\begin{tabular}{l|cc|cc|cc|cc}
\toprule
 & \multicolumn{2}{c|}{\textbf{Single Sentence}} & \multicolumn{2}{c|}{\textbf{Sentence Similarity}} & \multicolumn{2}{c|}{\textbf{Natural Language Inference}} & \multicolumn{2}{c}{\textbf{Commonsense Reasoning}} \\

&SST-2 & COLA & QQP & STS-B &  QNLI  &  RTE &  WinoGrande &  CSQA \\

\midrule
Human  
& 97.8 & 66.4 &  80.4  &92.7 & 91.2&93.6 & 94.0 &  89.0\\ 
\midrule
RoBERTa-Large 
& 96.2  & 64.9  &  91.8  &  92.2  &  94.4 &  84.3 & 78.1 & 72.1 \\
\midrule

``Task Difficulty" (\%)  &1.64 &  \textbf{2.26} & -14.18 &  \bf 0.54 &  -3.51 & \bf  9.94 &16.91  & \bf 18.99  \\
\bottomrule
\end{tabular}
\label{tab:across_task_difficulty}
}
\end{table*}
We acknowledge the limitations in our approach, such as when both humans and LLMs perform poorly on a task, potentially indicating high difficulty not reflected in our 'relative' gap metric. We hope our exploration can inspire more efforts in a more general and rigorous assessment of cross-task difficulty.

\section{Small-Magnitude Weights Contribute to Loss Basin Preservation} 
\label{appendix:lmc}
We also investigate the potential reasons behind the substantial performance drop resulting from the removal of small-magnitude weights on harder tasks. Our analysis revolves around the loss landscape, and we conjecture that small-magnitude weights play a significantly more crucial role in preserving the loss basin of the dense model on harder tasks compared to easier ones. Consequently, the absence of these small weights disrupts the optimal basin, leading to a considerable loss of performance.

To test our conjecture, we utilize the linear mode connectivity (LMC) metric proposed by ~\citep{frankle2020linear} between the solution produced by Sparse Transfer and that of Dense Transfer. Specifically, we perform linear interpolation between the fine-tuned model of Dense Transfer ($\bm{\theta}_d$) and the fine-tuned model of Sparse Transfer ($\bm{\theta}_s$), denoted as $\bm{\mathrm{\tilde\theta}} = \alpha \bm{\theta}_s + (1-\alpha) \bm{\theta}_d$. We conduct two sets of comparisons, namely QQP vs. STS-B and QNLI vs. RTE, and report the performance and loss in Figure~\ref{fig:LMC}. Our findings reveal that both sparse and dense models remain linearly connected, with minimal or no increase in loss barrier for easy tasks (i.e., QQP and RTE) when a certain portion of small-magnitude weights is removed. However, a significant increase in the loss barrier is observed when the same number of weights is removed for harder tasks. This observation strongly supports the concept of ``Junk DNA'', emphasizing the vital role of small-magnitude weights in ensuring that the fine-tuned model resides in the optimal basin. In contrast, the removal of these weights would lead to the destruction of this optimal basin, which is challenging to fix through fine-tuning,  causing a notable decline in performance. 


\begin{figure}[t]
\centering
\centering
    \subfigure{
        \includegraphics[width=0.5\textwidth]{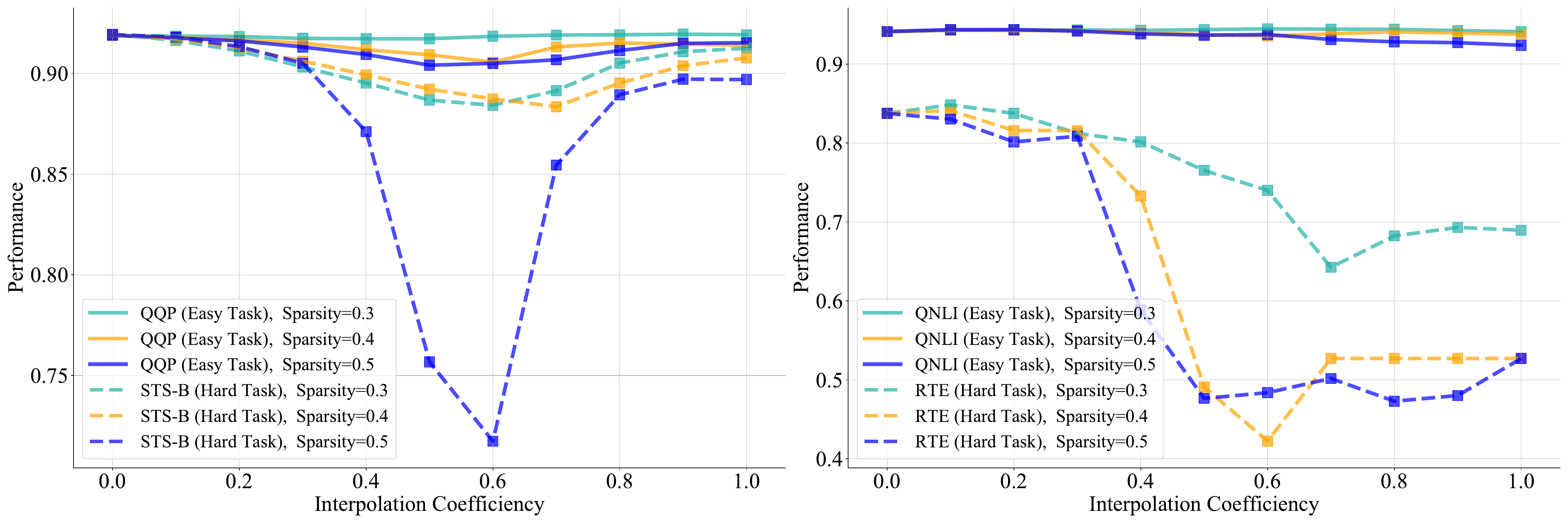}
    }

    \subfigure{
        \includegraphics[width=0.5\textwidth]{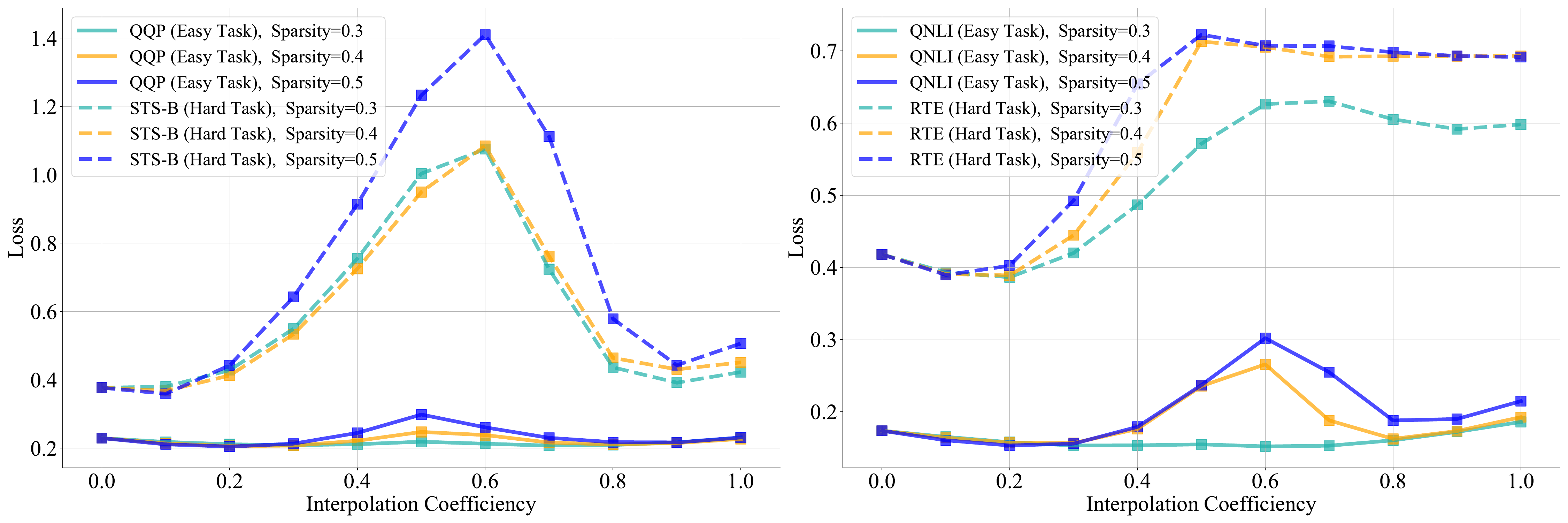}
    } 
\vspace{-1em}

\caption{Linear interpolation from the Dense Transfer (\textbf{Left}) model to its corresponding Sparse Transfer models (\textbf{Right}) on easy and harder tasks (in terms of across-task difficulty).}
\vspace{-2.5em}
\label{fig:LMC}
\end{figure}

\section{Dataset Details}
\label{appendix:dataset_details}
\subsection{Massive Multitask Language Understanding (MMLU)} MMLU \cite{hendrycks2020measuring} is a new benchmark designed to measure knowledge acquired during pretraining by evaluating models exclusively in zero-shot and few-shot settings. This makes the benchmark more challenging and more similar to how we evaluate humans. The benchmark covers 57 subjects across STEM, the humanities, the social sciences, and more. It ranges in difficulty from an elementary level to an advanced professional level, and it tests both world knowledge and problem solving ability. Subjects range from traditional areas, such as mathematics and history, to more specialized areas like law and ethics. The granularity and breadth of the subjects makes the benchmark ideal for identifying a model’s blind spots.

\subsection{Freebase Factoid Question Answering} FreebaseQA \cite{jiang-etal-2019-freebaseqa} is a data set for open-domain QA over the Freebase knowledge graph. The question-answer pairs in this data set are collected from various sources, including the TriviaQA data set and other trivia websites (QuizBalls, QuizZone, KnowQuiz), and are matched against Freebase to generate relevant subject-predicate-object triples that were further verified by human annotators. As all questions in FreebaseQA are composed independently for human contestants in various trivia-like competitions, this data set shows richer linguistic variation and complexity than existing QA data sets, making it a good test-bed for emerging KB-QA systems.

\subsection{TrviaQA Dataset} TriviaQA~\citep{joshi2017triviaqa} is a popular reading comprehension dataset which includes 95K question-answer pairs authored by trivia enthusiasts and independently gathered evidence documents, six per question on average, that provide high-quality distant supervision for answering the questions. TriviaQA consists of fairly complex compositional questions with considerable syntactic and lexical variability between questions and corresponding answer-evidence sentences, making it a challenging testbed for our evaluation.

\subsection{Glue Benchmark} General Language Understanding Evaluation (GLUE) benchmark \cite{wang2018glue} is a collection of nine natural language understanding tasks, including single-sentence tasks CoLA and SST-2, similarity and paraphrasing tasks MRPC, STS-B and QQP, and natural language inference tasks MNLI, QNLI, RTE and WNLI. GLUE is designed to favor and encourage models that share general linguistic knowledge across tasks.

\subsection{OpenbookQA Benchmark} OpenBookQA \cite{mihaylov2018can} is a new kind of question-answering dataset modeled after open book exams for assessing human understanding of a subject. It consists of 5,957 multiple-choice elementary-level science questions (4,957 train, 500 dev, 500 test), which probe the understanding of a small “book” of 1,326 core science facts and the application of these facts to novel situations. For training, the dataset includes a mapping from each question to the core science fact it was designed to probe. Answering OpenBookQA questions requires additional broad common knowledge, not contained in the book. The questions, by design, are answered incorrectly by both a retrieval-based algorithm and a word co-occurrence algorithm. Additionally, the dataset includes a collection of 5,167 crowd-sourced common knowledge facts, and an expanded version of the train/dev/test questions where each question is associated with its originating core fact, a human accuracy score, and a clarity score.

\subsection{HellaSwag} HellaSWAG \cite{zellers2019hellaswag} is a dataset for studying grounded commonsense inference. It consists of 70k multiple choice questions about grounded situations: each question comes from one of two domains - activitynet or wikihow - with four answer choices about what might happen next in the scene. The correct answer is the (real) sentence for the next event; the three incorrect answers are adversarially generated and human verified, so as to fool machines but not humans.

\section{Low-Rank Compression}\label{appendix:lrc}

An alternative way of compressing the model is reducing the rank of weight matrices by retaining the top $k$ components identified through Singular Value Decomposition (SVD)~\citep{lv2023lightformer,hajimolahoseini2021compressing,sharma2023truth}. In our study, we specifically modify the  rank of $
k$ using SVD decomposition across all linear layers in the transformer block of the Vicuna-7B model, ranging from 4000 to 100 ranks.. We then evaluate the compressed model's performance in two task difficulty settings: \ding{182} Setings 2: \textit{Varying the Option Count in Multiple-choice QA Setting} and Setings 3:\ding{183} \textit{Varying Context Length for Retrieval-Augmented QA}. 

\begin{figure}[h]
\centering
\centering
    \subfigure{
        \includegraphics[width=0.6\textwidth]{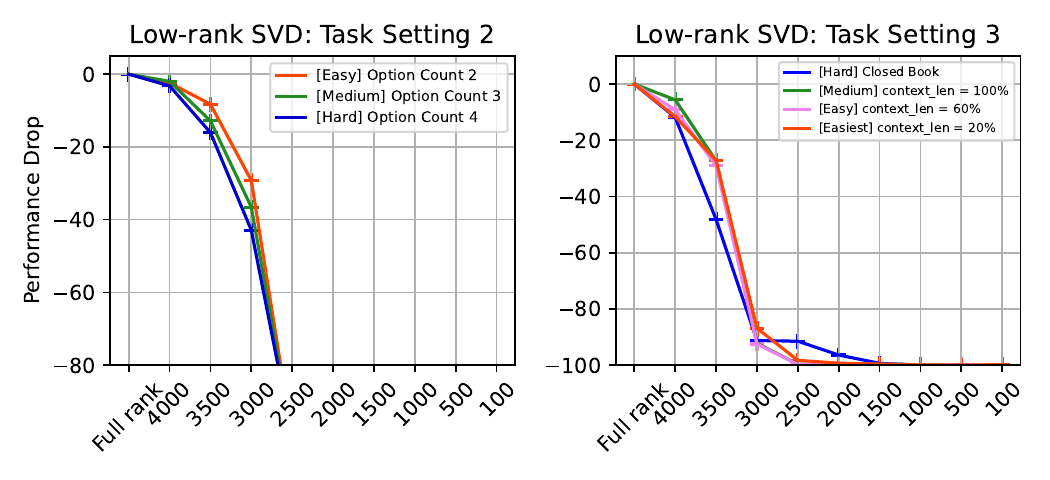}
    }

\caption{Low-Rank Compression using SVD.}
\label{fig:LMC}
\end{figure}

\begin{table}[h]
    \centering
    \caption{Low-Rank Compression using SVD.}
    \label{fig:LMC}
    \resizebox{0.55\textwidth}{!}{
    \begin{tabular}{l|cccc|ccc}
    \toprule
    & \multicolumn{4}{c|}{Context Length} & \multicolumn{3}{c}{Option Count} \\
    \cmidrule(r){2-5} \cmidrule(r){6-8}
         & Closed Book & 100\% & 60\% & 20\% & 2 & 3 & 4 \\
    \midrule
    Spearman's Rank & 0.899 & 0.942 & 0.905 & 0.912 & 0.933 & 0.956 & 0.956 \\
    \bottomrule
    \end{tabular}
    }
\end{table}


We noticed the concurrent work \cite{sharma2023truth} suggesting layer-selective low-rank compression of weights often improves LLM reasoning and generalization, without needing no re-training needed. We however note that requires careful selection of layer \& rank. In comparison, we apply ``uniform" SVD to all linear weight layers (except embedding), to make fair comparisons to pruning/quantization. We leave the examination of more sophisticated methods such as \cite{sharma2023truth} as future work.

\end{document}